\documentclass[conference]{IEEEtran}
\usepackage{times}

\usepackage[numbers,sort,compress]{natbib}
\usepackage{multicol}
\usepackage[bookmarks=true]{hyperref}
\usepackage{graphicx}
\usepackage{xcolor,colortbl}
\usepackage{silence}
\usepackage{caption}
\usepackage{subcaption}
\usepackage{balance}
\usepackage{makecell}
\usepackage{siunitx}
\usepackage{enumitem}
\usepackage{quoting}
\usepackage{booktabs}
\usepackage[noabbrev]{cleveref}

\newcommand{\say}[1]%
  {%
  \begin{quoting}[begintext=``, endtext='', vskip= 0.4ex, indentfirst=false]%
  #1%
  \end{quoting}%
  }%

\graphicspath{ {./figures/} }
\DeclareCaptionLabelFormat{andtable}{#1~#2  \&  \tablename~\thetable}

\begin{document}

% paper title

\title{Look Further: Socially-Compliant Navigation System in Residential Buildings}

% You will get a Paper-ID when submitting a pdf file to the conference system
% \author{Author Names Omitted for Anonymous Review. Paper-ID [1454]}

\author{
\IEEEauthorblockN{Akira Shiba$^*$, Marina Obata$^*$,  Nathan Kau$^*$, Zoltan Beck$^*$, Rishi Shah, Michael Sudano, Sabrina Lee}
\IEEEauthorblockA{Woven by Toyota, Inc., Japan\\
Email: [first].[last]@woven.toyota}
}

\maketitle
\def\thefootnote{*}\footnotetext{These authors contributed equally to this work}\def\thefootnote{\arabic{footnote}}

\begin{abstract}
The distance at which a mobile robot reacts to a person strongly impacts various qualities of the human-robot interaction.
In this paper, we focus on the navigation of a mobile delivery robot platform in a residential indoor hallway environment.
Social navigation methods typically focus on avoiding uncomfortable human-robot interactions, such as when a robot encroaches on someone's personal space.
Since personal space has been shown to be in the range of just a few meters, social navigation methods typically focus on deconflicting and resolving these short-range interactions. 
In this work, however, we demonstrate that by extending the reaction distance to over eight meters, far beyond the typical interaction distance, we can improve the human's perception of the robot's motion.
We introduce the Proactive Lane-Changing (PLC) motion pattern and a navigation system that leverages it to react to people at an increased distance.
This pattern consists of changing the robot's lateral position as it navigates down the hallway from the center to the side at an eight-meter distance from an oncoming person.

We conducted a user study with 42 participants to assess their impressions of the delivery robot based on three service objectives: safety, smoothness, and politeness.
In the straight hallway scenario (Frontal Approach), results showed significant improvement in each of these three objectives compared to typical motion patterns found in the literature: slowing down, stopping, and reactive collision avoidance in the proximity of a person.
In contrast, in the intersection (Blind Corner) scenarios, none of the approaches performed significantly better than any other, with participants having a diverse range of preferences among robot motion patterns.
\end{abstract}

\begin{IEEEkeywords}
logistics, robot, social navigation, long-range interaction, user study, proactive lane-changing
\end{IEEEkeywords}

\section{Introduction}
In recent years, an increasing number of mobile robot products have been introduced that are intended to operate alongside humans.
In a comprehensive survey paper of core challenges in social robot navigation, \citet{mavrogiannis2023core} highlights the need to organize and establish principled knowledge of human-robot social navigation and the importance of proxemics in Human-Robot Interaction. 
An extensive amount of research has been conducted on social navigation in unstructured environments such as large outdoor, public spaces.

Given that personal space is a few meters at most in radius \cite{LEICHTMANN2020101386}, in dynamic unstructured situations, reacting far ahead of a social encounter is not always effective since the situation is likely to change.
Such patterns can be observed in crowd simulations\cite{seitz2016cognitive}, where an interaction range of a couple meters produce realistic results.

Apart from crowd situations, there are various less dynamic contexts where people show a variety of behavior patterns.
In shops\cite{edirisinghe2024field} or museums\cite{hellou2022technical}, people move and stop between multiple landmarks, in order to search or browse.
Similarly, public hallways, such as university hallways (studied by \citet{macenski2020marathon}) may be used for several purposes: meeting people, rushing, walking, etc.

In the scope of this paper, we focus on shared hallways of residential apartment buildings.
We assume that people show a reduced number of behavior patterns compared to the examples above, spending most of their time getting to or from an apartment, causing lower variance in their motion preferences over time.
This allows a robot in such context to reliably plan their actions further ahead, benefiting from longer-range human sensing and reaction capabilities compared to existing literature in different contexts.

In this paper, we show how extended-range human-detection capability can result in a robot system that is able to navigate in an improved socially-compliant manner when compared to baseline systems with limited human-sensing capabilities.
Our main contributions are:
\begin{enumerate}
    \item We introduce a novel robot system with extended human-sensing capabilities that is able to react to people using a Proactive Lane-Changing (PLC) motion pattern 8 meters before an encounter.
    \item We evaluate this motion pattern against typical social navigation approaches with typical human-detection range as part of a user study with 42 participants in both \textit{Frontal Approach} and \textit{Blind Corner} encounters.
    \item Our findings show that the PLC motion pattern is rated significantly better with respect to our three service objectives (safety, smoothness, and politeness) in the \textit{Frontal Approach} situation, while there is no clear preference in the \textit{Blind Corner} encounter.
\end{enumerate}

\subsection*{Deployment Context}
\begin{figure}[ht]
    \centering
    \includegraphics[width=5.3cm]{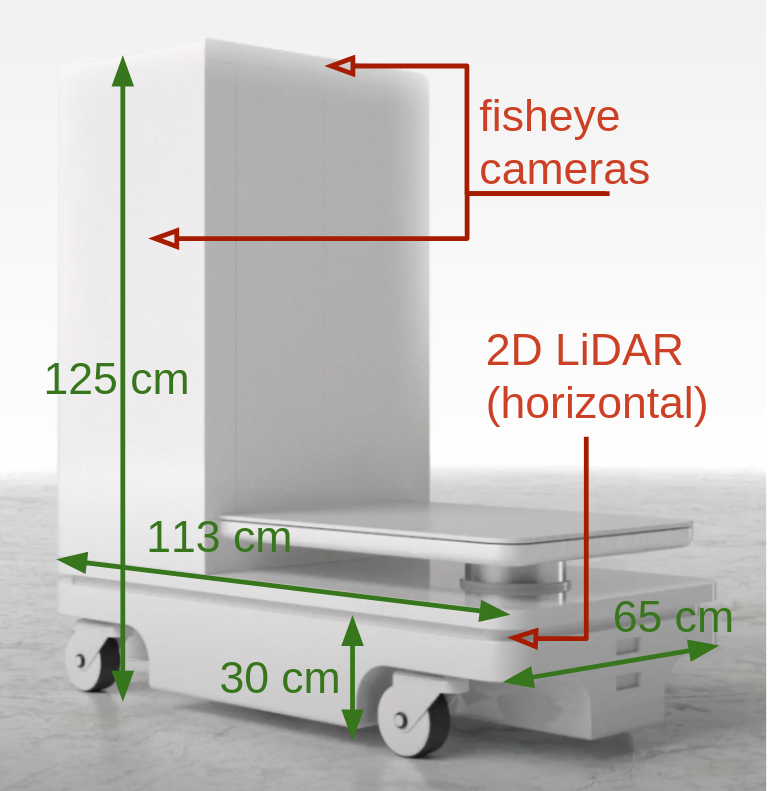}
    \caption{Smart Logistics Robot dimensions and HRI sensors}\label{fig:sl_robot}
\end{figure}

Our robot is intended for deployment in Toyota Woven City\cite{TWCweb}, located near Mount Fuji, Japan, and is launching in 2025.
Toyota Woven City has been designed as a `Test Course for Mobility', including the `mobility of goods' augmented by robotics, as part of this Smart Robotics Demonstration.
In more detail, a custom mobile robot platform (illustrated in Figure~\ref{fig:sl_robot}) performs package deliveries from a central logistics hub to both personal and shared storage facilities attached to residents’ apartments.

Residents in Toyota Woven City will actively participate in and test new services and technologies developed by inventors; as a result, they will play an essential part in the co-creation of projects through real-time feedback and engagement.
Such feedback provides a unique opportunity for the rapid improvement of the HRI behavior of the described Smart Logistics robot.

Operating in and around residential apartment buildings, the robot will often encounter people.
Though these encounters will be limited interactions in building hallways, a human-centered approach should be used when designing and implementing the robot navigation system.
In order to achieve this, the Smart Logistics service has three objectives: (1) safety, (2) smoothness, and (3) politeness.
These are the main metrics against which we will evaluate our system throughout the rest of the document.

\section{Related Work}

\subsection*{Implicit Social Navigation}
Methods that implicitly encode social behaviors include well-studied general-purpose techniques such as DWA \cite{DWA}, MPPI \cite{MPPI_7487277}, ORCA \cite{ORCA_van2011reciprocal}, and the timed elastic band \cite{TEB} which primarily focus on navigation in unstructured environments and crowds and generally perform short-range obstacle avoidance. These methods are primarily designed such that socially compliant navigation emerges, rather than it being built in inherently. On the other hand, data-driven methods include inverse reinforcement learning for informing navigation based on human goals \cite{IRL_henry2010learning, IRL_kretzschmar2016socially, IRL_ziebart2009planning}. Another prominent area of data-driven methods is behavior cloning \cite{shah2023vint, xiao2022learning}. In \cite{shah2023vint}, a goal-image-conditioned model is trained on diverse mobile robot driving datasets, including both indoor and outdoor environments and shows dynamic pedestrian avoidance for short-range encounters. However, social behaviors beyond pedestrian avoidance are not shown and human preferences are not studied. Using a learned MPC model, \citet{xiao2022learning} trains a policy specifically on expert demonstrations of socially compliant movement in indoor environments. In a qualitative study, the learned policy is rated more highly than a baseline non-learned navigation system in a blind corner and pedestrian obstruction situation. In both works, the learned models are tested in short-range interactions, such as blind corners, pedestrian obstruction, and frontal encounters. Instead of using a data-driven approach, we explicitly develop a simple pro-social movement pattern so that the robot can react consistently, predictably, and from a further distance from the person, in order to deconflict the interaction early on.

\subsection*{Explicit Social Navigation}
Robots with navigation systems containing explicitly defined pro-social patterns have been found to significantly improve human perceptions of encounters.
To solve the freezing robot problem, where robots freeze due to a conservative policy, unable to move in a dynamic environment, \citet{kamezaki2020preliminary} showed improved psychological reactions compared to baselines in an indoor environment that used multi-modal inducement to instigate a person to move out of the way of a path-obstructed robot.
In another study in an indoor environment, \citet{singh2023behavior} experimented with different behaviors akin to the speeds and motion conditions of a delivery robot in a hotel corridor.
They concluded, using a non-learning-based approach, that a robot with social behavior is more acceptable in a human-populated environment.
\citet{physiological} found that a robot using a Social Forces Model \cite{SFM} to avoid people was evaluated qualitatively as more comfortable and less threatening than a non-social baseline.

Regarding the comparison of various passing behaviors from the bystander's perspective, \citet{5453230} tested four variations with stopping, slowing, neutral, and fast passing combined with four different robot types via a web survey with video clips. They highlighted that the stop behavior was rated as the most polite regarding the robot's adherence to social etiquette, as well as the most trusted behavior, and in both cases, the fast passing behavior was the lowest rated one. 

In a similar vein, \citet{geldenbott2024legible} and \citet{companion_5326271} introduce planning methods that make the robot proactively move to the side to make room for pedestrians. \citet{geldenbott2024legible} accomplishes this by incorporating a markup term in an MPC formulation that encourages the robot to make larger control inputs earlier rather than later. \citet{companion_5326271} proposes a planner which encourages the robot to pass people on the culturally normative side by adding an asymmetric cost field around people at a global planner level. In a follow-up user study \cite{kirby2010social}, the authors find significant effects on some human ratings. While our proposed system also augments the global planner with additional cost terms to achieve PLC, we focus on enacting our pro-social robot motion from a further distance (\SI{8}{\metre}) to deconflict the situation at an earlier point. Additionally, the authors of \cite{geldenbott2024legible} do not perform a real-life user study, and \cite{companion_5326271} does not detect people in real time in its experiments or compare against a non-social baseline; in contrast, our system detects people online and we compare PLC against several social and non-social baselines.

\subsection*{Person Detection}
Person tracking is a well studied field\cite{hasan_lidar-based_2022} with methods including camera-only and LiDAR-only approaches, as well as approaches that do and do not use learning. Systems that include 3D LiDAR achieve state-of-the-art performance, but due to cost considerations our robot only features 2D LiDAR. To overcome the limitations of 2D LiDAR, and to increase versatility compared to heuristic-based approaches, our work adopts a learning-based person detector based upon work by \citet{person_detector} that combines multi-view camera images and 2D LiDAR data to localize people around the robot in 3D. 

\subsection*{Wizard of Oz}
We began our study by conducting an exploratory pilot study using the Wizard of Oz framework to quickly and effectively\cite{exploratory} inform our research questions and the design of our navigation system.
The Wizard of Oz framework has been used extensively for a variety of robot systems in order to quickly understand and verify human preferences, such as those around verbal and non-verbal communication, robot movements, and indoor navigation \cite{gibert2013makes, 8172365, 8172374, exploratory, 10.1145/2701973.2702060}. This framework has also been used in the design of autonomous vehicle systems \cite{10.1145/3461778.3462056, 8569486, driving28548, robotaxi_wooz}. We use the Wizard of Oz framework in a pilot study to attain qualitative insights that form the basis of our proposed navigation system. To our knowledge, our pilot study is the first use of this framework for indoor navigation of a delivery robot.

\section{Pilot Study}
We conducted a Wizard of Oz pilot study to understand the ideal socially-compliant motion pattern of our mobile robot when passing a person in a straight corridor.
The pilot study was conducted with 5 female and 5 male participants (age group mixed), and the following motion patterns.
\begin{enumerate}[label=Pattern \arabic*, leftmargin=*]
    \item \label{pat:1}The robot stops at a specific distance from the person.
    \item \label{pat:2}The robot starts on the side of the corridor and continues down with constant nominal velocity.
    \item \label{pat:3}The robot starts in the middle of the corridor and then moves to the side of the corridor at a given distance from the person.
\end{enumerate}

The reaction distances for \ref{pat:1} and \ref{pat:3} were tested at \SIlist{6; 8; 10}{\metre} that was determined using our initial findings when testing reaction distances ranging from 2 to \SI{10}{\metre} with \SI{2}{\metre} increments. At distances of \SIlist{2; 4}{\metre}, the robot's abrupt avoidance behavior negatively impacted perceived safety. As can be seen in Figure~\ref{fig:globalpath}, the distance between the robot and the participant decreases by about \SI{5}{\metre} during autonomous operation.
 
The study was reviewed by internal privacy and safety teams in advance.
As an experimental protocol, participants were briefed on the study flow, data collection, and safety precautions, and were then asked to imagine using the delivery service.
After completing each task, primary qualitative feedback and supplemental quantitative feedback was recorded on a 5-point Likert scale.

The qualitative findings from this study suggested the following:
\begin{itemize}
    \item \ref{pat:1} was generally evaluated as the lowest,
    \item Some participants expressed anxiety from \ref{pat:2}, stating that it was unclear if the robot might suddenly change its position,
    \item Participants preferred the sideways motion in \ref{pat:3}; however, they found it uncomfortably close when the reaction distance was \SI{6}{\metre} and some found it confusing when it was \SI{10}{\metre}.
    
\end{itemize}
As it was the best-performing motion pattern, we decided to explore \ref{pat:3} and implemented PLC starting at a distance of \SI{8}{\metre} from a person.

\section{Navigation System}
\begin{figure}[ht]
    \centering
    \includegraphics[width=\columnwidth]{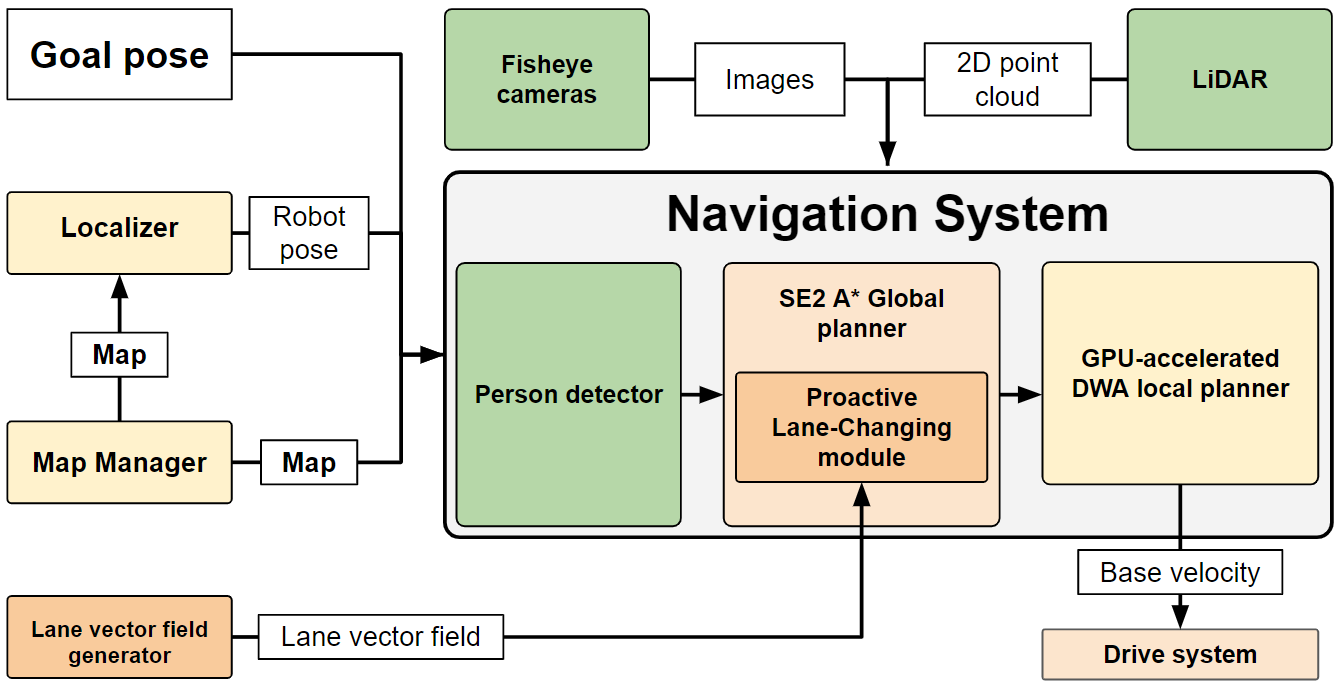}
    \caption{System diagram of proposed navigation system. Images from fisheye cameras and point clouds from LiDAR are input to a navigation system which uses the data to perform person detection. Person detections are input to the global planner which also uses a pre-computed lane vector field to proactively change lanes in the presence of people.}
    \label{fig:system_diagram}
\end{figure}

The navigation system (Figure \ref{fig:system_diagram}) that implements PLC is comprised of three main components:
(1)~an enhanced, data-driven module that detects the 3D position of people in the environment,
(2)~a Bayes filter to provide temporal consistency to the detections and estimate whether a person is oncoming with respect to the robot, and
(3)~a motion planner that augments a conventional navigation stack, consisting of local and global planners, along with our novel PLC system. 

\subsection*{Robot Platform}
As shown in Figure \ref{fig:sl_robot}, the mobile robot used in the study has a width of \SI{65}{\centi\meter}, a length of \SI{113}{\centi\meter}, and a height of \SI{125}{\centi\meter}.
The robot has an omnidirectional drive base consisting of a combination of powered caster wheels, passive omnidirectional wheels, and idler casters. To manipulate packages, the robot has a table that can be raised and lowered and interfaces with an intelligent shelving system located at the rear of the robot.
The robot features six short-range depth cameras (used only for collision avoidance); four fisheye cameras located on the front, left, right, and back sides of the rear shelving tower; and two horizontal 2D LiDARs attached to the base of the robot at a height of \SI{30}{\centi\meter}.

Under nominal conditions, the robot travels at \SI{0.4}{\meter/\second} and has a maximum lateral velocity of \SI{0.2}{\meter/\second}. 

\subsection*{Person Detector}

% \textbf{Needs editing.} 

The person detector fuses data from multiple 2D LiDARs and fisheye cameras in order to identify and locate people around the robot in a approach based on \citet{person_detector}. To increase the robustness and accuracy, we use modern pre-trained networks and architectures, use additional filtering in the data autolabelling pipeline, and integrate support for fisheye cameras. The Images from the four fisheye cameras mounted around the outside of the robot are batched and input to the Yolo~v7 object detector \cite{wang2022yolov7trainablebagoffreebiessets} in order to obtain person bounding boxes.
Scan points from two 2D LiDARs are then projected into each of the bounding boxes using calibrated double sphere distortion models \cite{usenko2018double}. To distinguish the projected points within the bounding boxes between those that correspond to people and those that do not (for example, those of the foreground or background), we train and use a lightweight transformer \cite{vaswani2023attentionneed} seq2seq model that classifies each point as either ``person'' or ``not person''. Finally, for each bounding box, the median coordinates of the ``person'' points within a bounding box is output as the person's position.

The transformer model used to classify LiDAR points as ``person'' or ``not person'' is trained with a self-supervised auto-labeling pipeline. Given LiDAR and fisheye image data collected by driving the robot in an indoor environment, the auto-labeler classifies LiDAR points by first using SegmentAnything \cite{kirillov2023segment} to filter out points that do not fall within the segmentation of the person in each bounding box. While this filtering is sufficient for \citet{person_detector}, because our cameras and LiDARs are mounted relatively far apart on the robot, \cite{person_detector} by itself produces a higher rate of false-positives. To remedy this,  MMPose \cite{mmpose2020} is used to reject any points that do not lie between the ankle and the knee of the person. One hour's worth of data collected in an office environment was sufficient to generate enough data to train the transformer model and reach AP@\SI{0.3}{\metre} of 95.0\% and AP@\SI{0.5}{\metre} of 98.8\% on our validation dataset.

\subsection*{Bayes Filter}
To mitigate cases where the YOLOv7 object detector drops detections from frame to frame, we temporally smooth the detections, using a binary Bayes filter to estimate the binary state $x$ of whether any person is in front of the robot and within a given distance $d_{activation}$, set to \SI{8}{\meter}. The person detector provides the state estimate and we tune the transition and measurement matrices to achieve desired robustness against dropped detections.

\subsection*{Motion Planner \& PLC}
The motion planner combines (1) a conventional navigation stack based on \cite{macenski2020marathon} that uses an SE2 A* global planner to generate global paths and a GPU-accelerated DWA \cite{DWA} local planner to generate vehicle velocities with (2) our PLC system. 

If the Bayes filter belief $x$ is less than a threshold $\epsilon$ then the robot performs navigation with no explicit social behavior. However, if the belief $x$ exceeds the threshold $\epsilon$, then the PLC system is activated, moving the robot to the left side, which is the culturally normative side in Japan.
To achieve this behavior, the PLC system uses a pre-computed lane map of the indoor environment, represented as a discrete vector field where each \SI{5}{\centi\meter} x \SI{5}{\centi\meter} cell is associated with a 2D lane vector.

This vector field is generated automatically from the binary rasterized representation of the environment that is used for obstacle avoidance. 
Once activated, the PLC module sets a non-zero weight to an additional lane following term in the edge cost computation in the A* global planner to encourage the robot to travel in the specified lane at a specified $d_{wall}$ distance from the hallway wall. The lane following term is formulated as:
\[c_{lanes} = 0.5 - 0.5 \mathbf{v} \cdot \mathbf{\hat{d}}\]
where $\mathbf{v}$ is the 2D lane vector at the edge’s parent node in x and y coordinates (with maximum magnitude of 1.0), and $\mathbf{\hat{d}}$ is the 2D unit vector representing the edge direction. The advantage of treating the lane following term as an additional cost is that it can be combined gracefully with other typical reward terms such as facing forward, giving buffer zones around people, or avoiding high cost areas like inside corners of hallway corners \cite{macenski2020marathon}.

\begin{figure}[ht]
    \centering
    \includegraphics[width=\columnwidth]{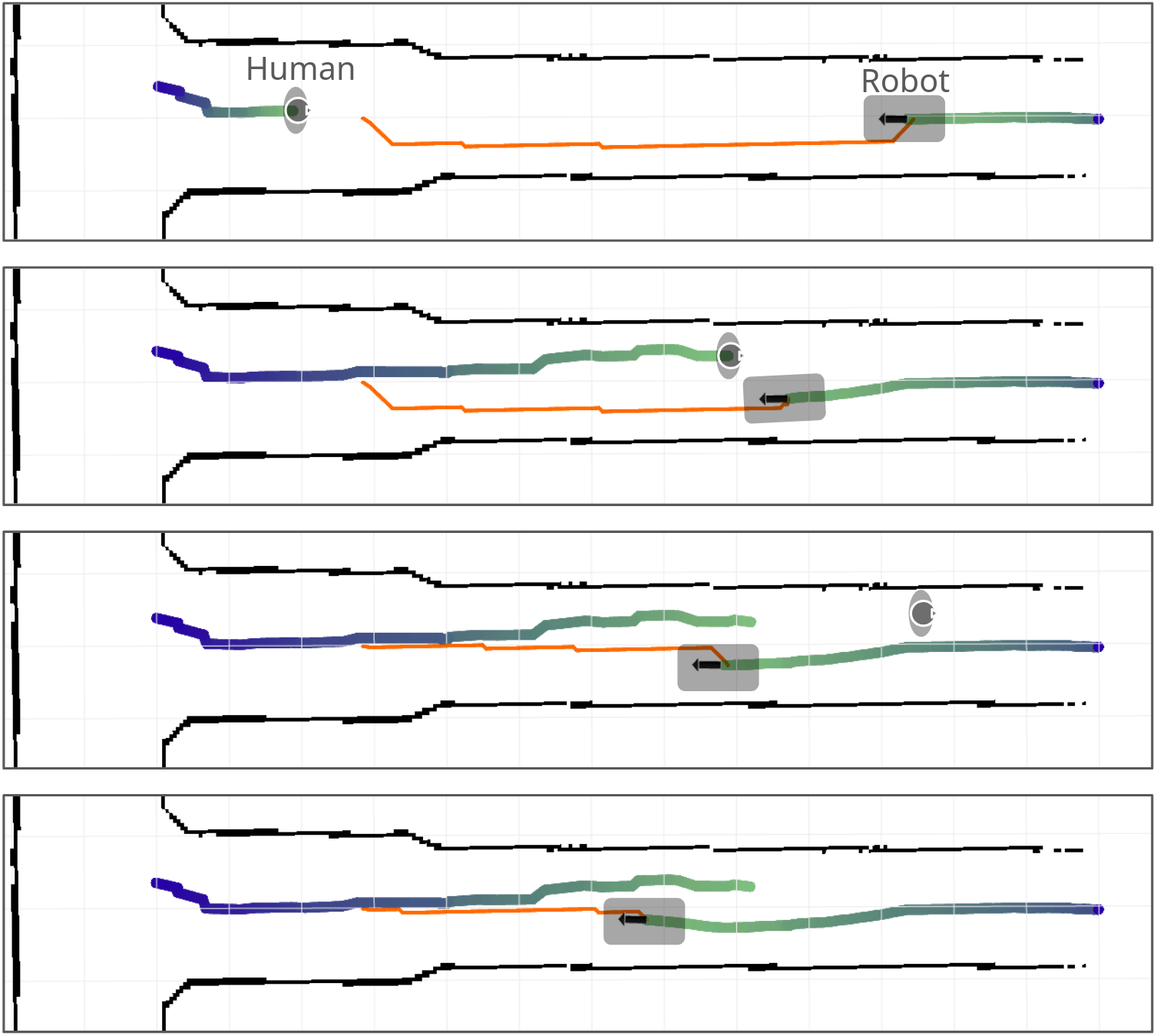}
    \caption{PLC behavior during hallway encounter.
    The top image shows the robot and the human at the start of the interaction, and the following ones shows the state as the time progresses.
    The current poses of the robot and the human are marked in gray, their past trajectories with thick green-blue lines and the robot's planned trajectory is displayed with a thin orange line.
    The trajectory of the person stops when they move past the camera of the robot, as the robot no longer detects their position. The grid resolution is \SI{1}{\metre}.}
    \label{fig:globalpath}
\end{figure}

Figure \ref{fig:globalpath} illustrates how the robot's path biases towards the lane when a person is detected in a real environment both for the A* trajectory plan and the executed robot trajectory.

\section{User Study}

% \subsection*{Motivation and Research Question}
This study was conducted to evaluate the navigation system with PLC and to ensure that the robot can meet our service objectives, safety, smoothness, and politeness, before deployment as part of the Smart Logistics service in Toyota Woven City.

The study was designed to answer the following research questions:

\textbf{RQ1:} Does the social navigation system with PLC behavior and long-distance person detection improve impressions of the delivery robot compared to baseline robot navigation systems?

\textbf{RQ2:} Does the social navigation system with PLC improve impressions when implemented in residential settings with limited visibility?

\subsection*{Settings of the corridor and measured behaviors}

From \cite{Francis2023PrinciplesAG}, two of the social navigation scenarios were selected: Scenario A - \textit{Frontal Approach} and Scenario B - \textit{Blind Corner}. Scenario A was selected due to it being the most common robot-human interaction within the residential buildings. For the purposes of calculating passable space \SI{1.6}{\metre}, the minimum width for a hallway with residential units on both sides\cite{e-gov} was used. Scenario B was selected as the most challenging situation that could occur within the residential buildings. Where corners at intersections create a situation with reduced reaction time due to limited visibility. From \cite{Francis2023PrinciplesAG}, \textit{Doorway}, \textit{Interpersonal}, and \textit{Specialized} scenarios were determined to not be situations that could occur within the residential buildings and were excluded. For the initial study, it was assumed that the hallways would not be crowded, and the \textit{Crowd} scenarios were also excluded.

To address RQ1, the PLC behavior (Task~4) was compared with three different baseline control behaviors: (Task~1) nominally constant velocity with short-range collision avoidance, (Task~2) stopping when within \SI{1.2}{\meter} of a person, and (Task~3) slowing when within \SI{1.2}{\meter} of a person with short-range collision avoidance. In Tasks~2 and 3, the robot stops or slows down \SI{1.2}{\meter} away from the participants; given the social distance range is \SIrange{1.2}{3.6}{\metre} and the comfortable distance with a neutral face robot was \SI{1.19}{\meter} found in \cite{distance, 10.1007/978-3-319-50115-4_69}.
We conducted the comparative evaluation with short-distance collision avoidance combined with stopping and slowing down behaviors as previous studies have demonstrated them to be preferred behaviors for short-range interaction \cite{singh2023behavior, 5453230}.
While other, more elaborate baselines exist, our assumption is that our collision avoidance approach utilizing DWA represents other state-of-the-art approaches (MPPI, SFM, ORCA, learned MPC, etc.) when running on our low velocity robot in a narrow hallway as the action space is very limited.

\begin{table}[ht]
    \centering
    \begin{tabular}{@{}llll@{}}
        \toprule
         & \multicolumn{2}{r}{Scenarios} \\
       Task & Behavior & A: Frontal Approach & B: Blind Corner \\  \midrule
         1 & Constant & Left side, $d_1$=\SI{0.65}{\metre} & Center, A--B\\
        2 & Stop & Left side, $d_1$=\SI{0.65}{\metre} & Center, A--B \\
        3 & Slow down & Left side, $d_1$=\SI{0.65}{\metre} & Center, A--B \\
        4 & PLC & Center \textrightarrow ~Left, $d_2$=\SI{0.45}{\metre} & Center \textrightarrow ~Left, A--B\\
        $4'$ & PLC & --- & Center \textrightarrow ~Left, A'--B \\ \bottomrule
    \end{tabular}
    \caption{Combinations of scenarios and tasks and the robot running position for each.
    The running position from the wall, $d_1$, was determined to maintain just enough walking space for the participants (\SI{0.7}{\meter}: covers 95th percent of the forearm-forearm breadth of the 40-year-old American male\cite{NASA}), but still leave distance from the other wall in case someone exits their apartment on the left side of the corridor.
    On the other hand, $d_2$ for PLC results in a running position closer to the wall as the robot only approaches the side during an encounter.}
    \label{tab:tasks}
\end{table}

\begin{figure*}[ht]
    \centering
    \includegraphics[keepaspectratio, width=0.8\linewidth]{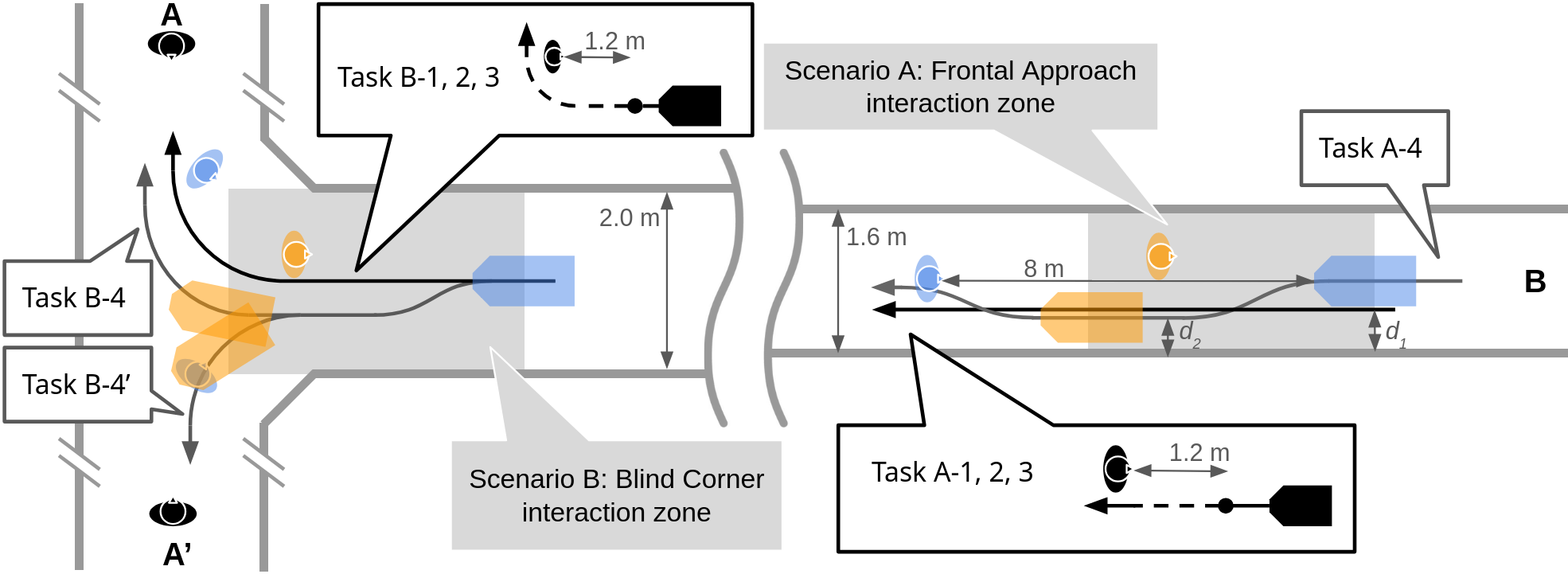}
    \caption{Study layout in the two scenarios: \textit{Frontal Approach} (right) and \textit{Blind Corner} (left). Three of the 4 tasks (1, 2, and 3) follow the trajectory marked with black, while task 4 (and $4'$) follow the one in gray (more details in Table~\ref{tab:tasks}). Blue color represents the robot and human at the start of their interaction, while orange at the end of their interaction for Task~4 (and $4'$).}\label{fig:layout}
\end{figure*}

RQ2 was addressed by testing whether the PLC behavior can be utilized in Scenario B - \textit{Blind Corner} where the visibility is naturally limited by the walls.
In this scenario, the first three tasks were only included in one of the two corner directions due to their symmetry, while the PLC behavior was included in both turn direction settings (Tasks 4 and $4'$) as PLC will change the running position to the left regardless of the turn direction resulting in asymmetrical behavior. 

\subsection*{Participants}
For confidentiality reasons, the participants were internal company employees
(\textit{N}=42, 24 men and 18 women; 4 in their 20~s, 17 in their 30~s, 16 in their 40~s and 5 in their 50~s).
We screened out those who have been living in Japan for less than one year to ensure that the participants were well-adapted to the local traffic rules and those who had any traumatizing experience with robots to protect the participants’ psychological well-being and baseline emotional neutrality. 
Prior to participation, all participants were informed about the experimental procedure and methods for data acquisition with video instruction and signed informed consent documents. In lieu of an IRB process,
% Woven By Toyota's
corporate privacy and safety teams conducted internal reviews of the study.

\subsection*{Experimental Protocol}
Participants were first introduced to the study and the Smart Logistics’ service provision, followed by consent procedures.
Then, participants were asked to perform nine tasks consisting of the four (or five) robot tasks in two scenarios (see Table~\ref{tab:tasks}).
The order of the tasks was randomized within each scenario for each participant group (of a maximum of 3 participants).
Before starting the tasks, each participant's walking speed was measured as a control for comparison.
The participants were instructed to imagine themselves living in the city as residents as if they were about to go grocery shopping or go to work and walk as they naturally do in their daily lives.

In each task, the participant walked one way through the mock-up hallway from point A to point B (see Figure~\ref{fig:layout}) over a total distance of \SI{15}{\meter}, interacting with the robot along the way. Participants were instructed to start walking when they heard a doorbell sound, at which point the research staff started a stopwatch. The stopwatch was stopped when the participant pressed the button at the destination (point B). The robot was autonomously driven, but a robot operator monitored the robot's actions at all times in the same room to ensure safety. After completing each task, participants were asked to fill out the post-task questionnaire. After completion of all tasks and scenarios, participants were asked to answer a few open-ended questions in a face-to-face interview. 

\begin{figure}[ht]
    \centering
    \begin{subfigure}[b]{0.48\columnwidth}
        \centering
        \includegraphics[width=\textwidth]{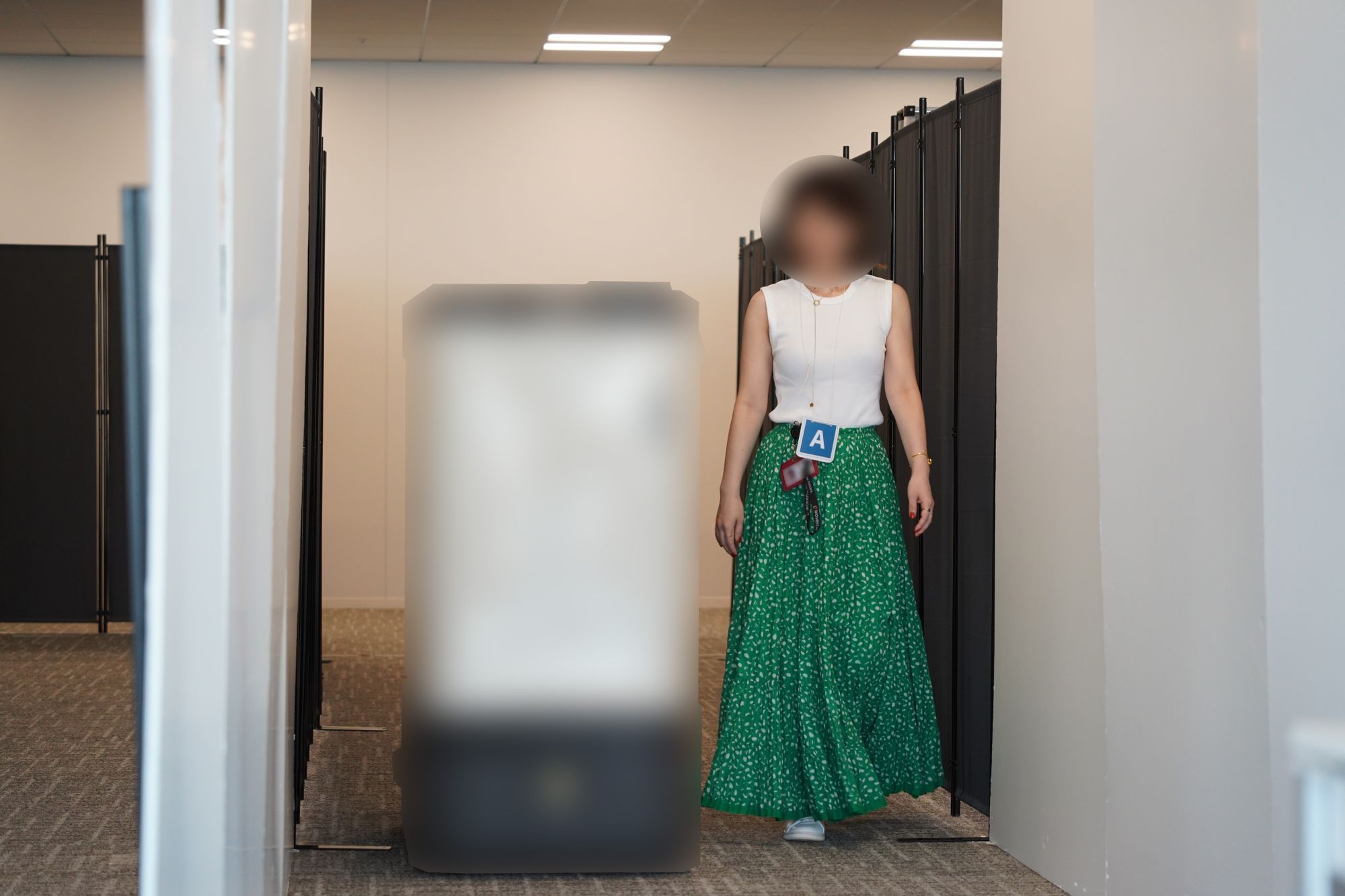}
        \caption{\textit{Frontal Approach} interaction}
        \label{fig:photo-hallway}
    \end{subfigure}
    \hfill
    \begin{subfigure}[b]{0.48\columnwidth}
        \centering
        \includegraphics[width=\textwidth]{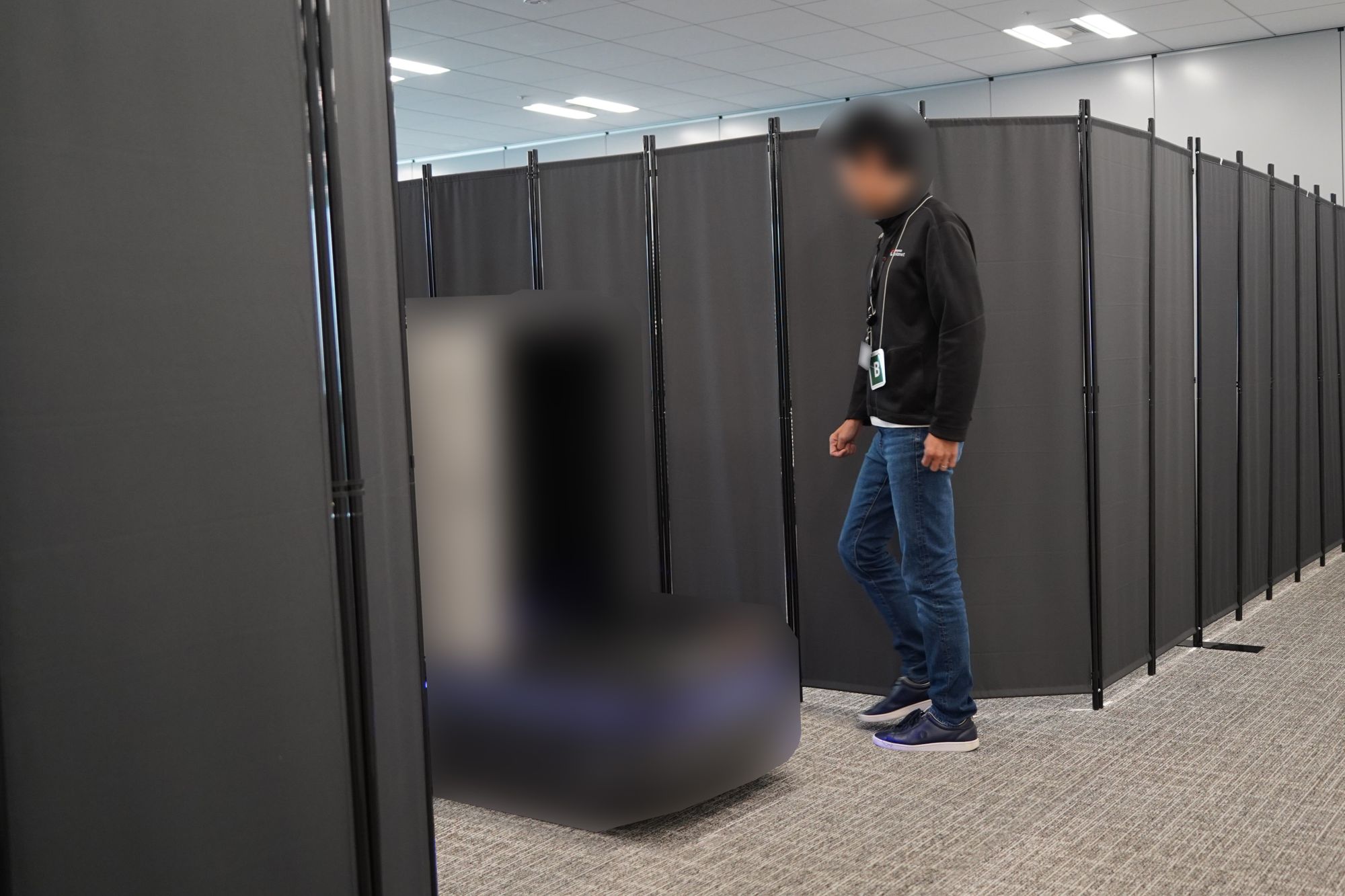}
        \caption{\textit{Blind Corner} interaction}
        \label{fig:photo-intersection}
    \end{subfigure}
    \hfill
    \begin{subfigure}[b]{0.48\columnwidth}
        \centering
        \includegraphics[width=\textwidth]{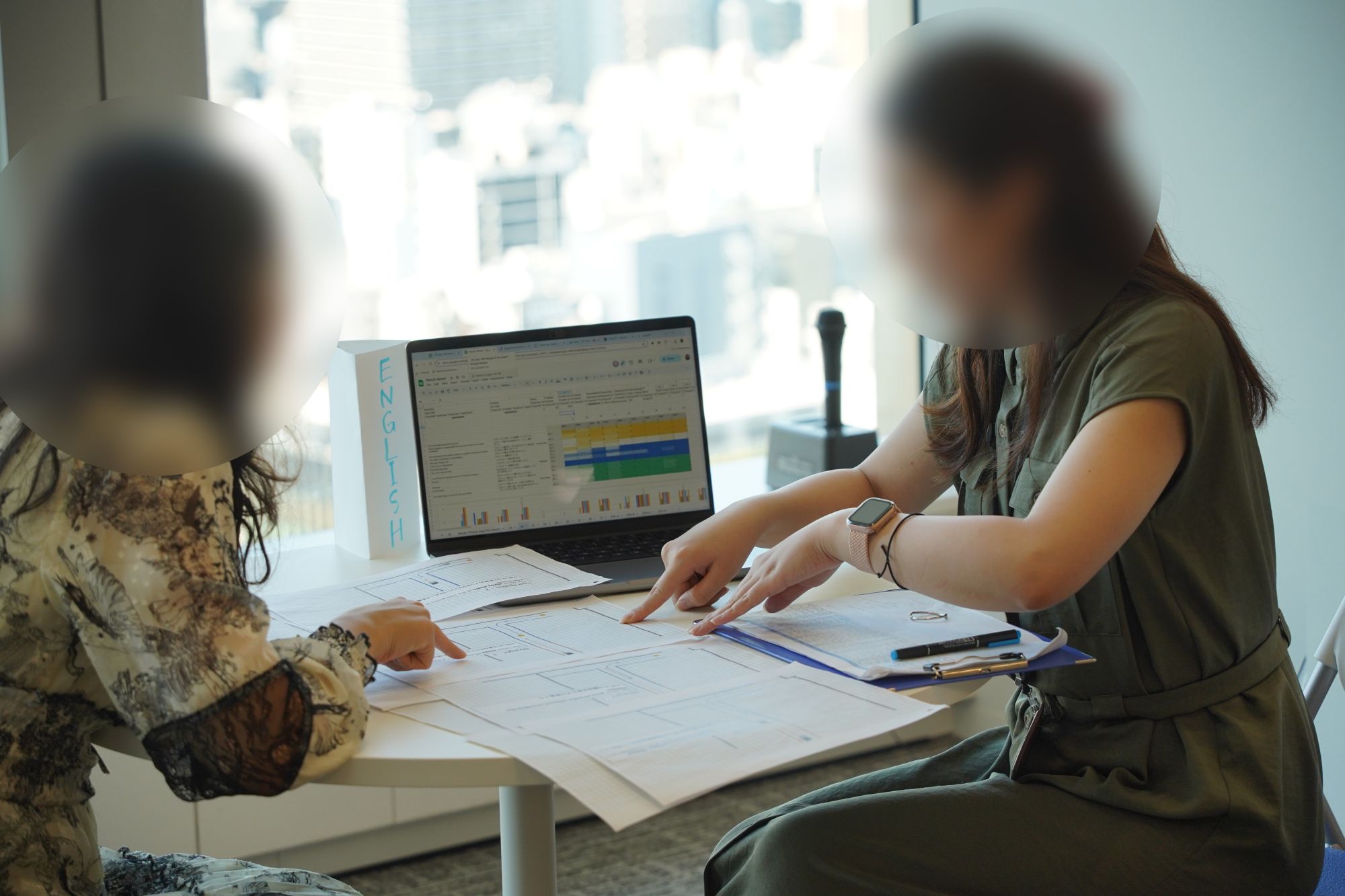}
        \caption{Follow-up interview}
        \label{fig:photo-interview}
    \end{subfigure}
    \caption{Photos from the research setting. Robot and faces blurred due to confidentiality.}\label{fig:photos}
\end{figure}

To minimize influences within a participant group, only one person encountered the robot at a time while the remaining participants completed questionnaires.
The encounters occurred behind 2-meter-high walls, which effectively prevented visibility and minimized communication.
Some photos of the study (with the participants and the robot blurred for confidentiality reasons) can be seen in Figure~\ref{fig:photos}.

\subsection*{Measurement Variables}
To measure our service objectives of (1) safety, (2) smoothness, and (3) politeness, subjective evaluations were conducted based on \cite{s20236822, lee2011effect, AKALIN2022102744}(see Table~\ref{tab:measurement-vars}). Additionally, an objective evaluation of each participant's movement efficiency, based on \cite{Francis2023PrinciplesAG}, was implemented to understand possible trade-offs with the subjective measures. 
Subjective evaluations were measured by a Likert scale \cite{7Likert} that has been recommended to evaluate social robot navigation \cite{Francis2023PrinciplesAG}.
As we aimed to measure more distinct preferences, a seven-point Likert scale was applied\cite{article_Likert}. 
The objective evaluation was conducted by measuring the efficiency of participants’ travel between points A and B. As each participant evaluated all nine scenarios, statistical analysis of evaluation metrics was conducted via within-subjects comparison.
The follow-up interview aimed to understand whether the participants perceived any difference between scenarios, the most and least preferred interaction, and the safest, smoothest, most polite, and most efficient interaction, followed by their reasoning \cite{article_kumar}.

\begin{table}[ht]
\begin{tabular}{|l|l|l|l|}
\hline
\makecell{Evaluation\\metrics} & \textbf{1 Safety} & \textbf{2 Smoothness} & \textbf{3 Politeness} \\ \hline
\makecell{Questionnaire\\items} &
\makecell{Secure,\\[3pt]Relaxed,\\[3pt]Comfortable,\\[3pt]In control of\\ own safety} &
\makecell{Takes people into\\account,\\[3pt]Navigation is\\appropriate from our\\social convention,\\[3pt]Similar to a person in\\the same situation,\\[3pt]Unpredictable,\\[3pt]Disturbs me} &
\makecell{Responsible,\\[3pt]Polite,\\[3pt]Sincere,\\[3pt]Want to use\\ the service}\\ \hline
\end{tabular}
\caption{Items for subjective measurement}
\label{tab:measurement-vars}
\end{table}

\begin{figure*}[!t]
    \centering
    \includegraphics[width=0.85\linewidth]{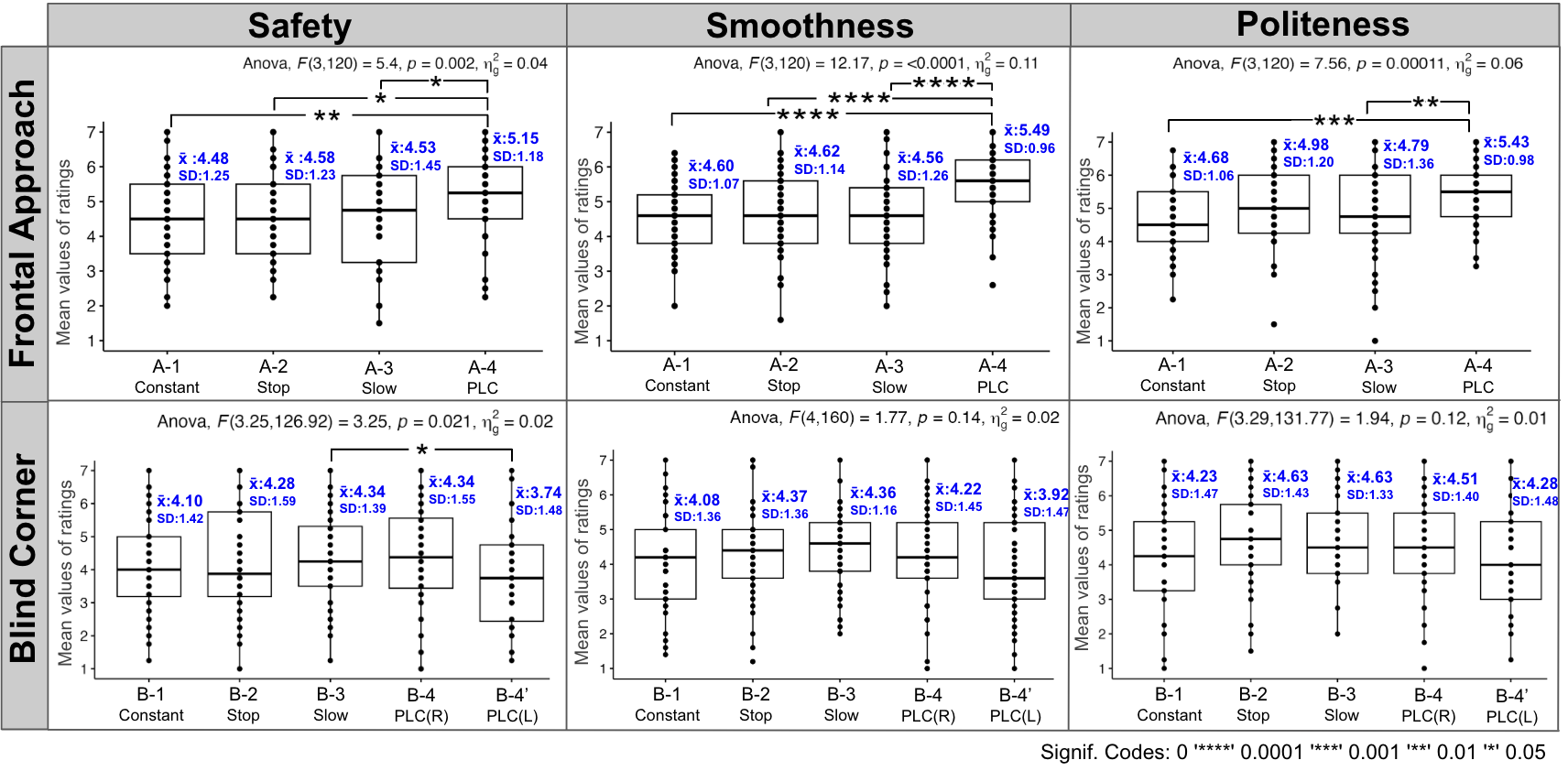}
    \caption{Subjective evaluation on safety, smoothness, and politeness across \textit{Frontal Approach} and \textit{Blind Corner} scenarios.} %PLC was evaluated as significantly safer and smoother than all other tested behaviors in the \textit{Frontal Approach} scenario. It was also evaluated as significantly more polite compared to slowing down and constant behaviors. In the \textit{Blind Corner} scenario, significant within-subjects difference was found only for the safety category.}%
    \label{fig:result-individual}
\end{figure*}
    
%Objective evaluation on efficiency;

\begin{figure}[ht!]
    \centering
    \includegraphics[width=0.62\columnwidth]{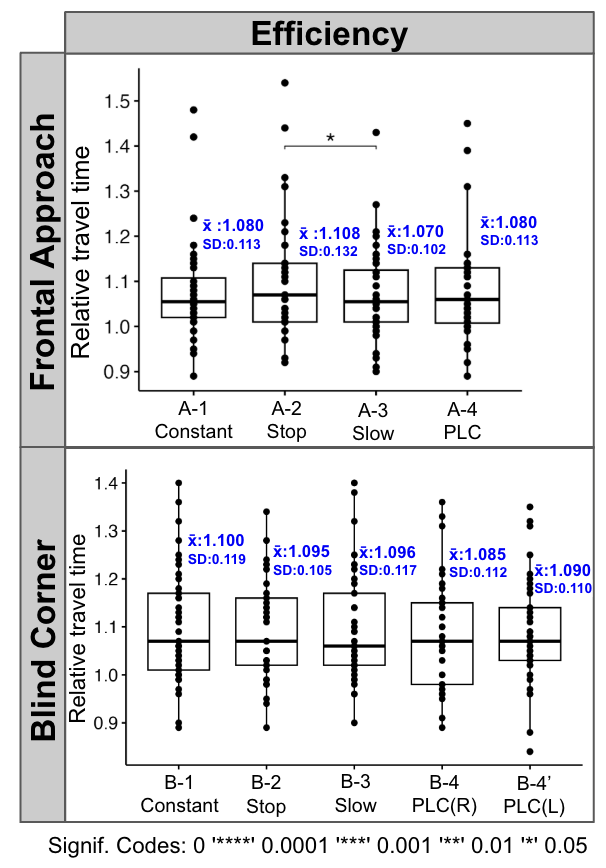}
    \caption{Comparison of movement efficiency across four behaviors, measured by relative travel time compared to a per-particpant control walking speed.}
    \label{fig:eff_total score}
\end{figure}

\section{Study Results}
The quantitative results are displayed in Figure~\ref{fig:result-individual} and Figure~\ref{fig:eff_total score}. Out of the forty-two participants, one participant who took a clearly provocative stance toward the robot was removed from the entire analysis. There were two instances of missing walking speed data, one from \textit{Frontal Approach} and one from the \textit{Blind Corner} scenario, efficiency analysis was not performed for that task in both instances. 

\subsection*{I. Frontal Approach}
The subjective results, shown in \Cref{fig:result-individual}, come from the questionnaire items measured by a 7-point Likert scale that have been averaged within each evaluation category (safety, smoothness, and politeness). Repeated measures ANOVA tests\cite{ANOVA} demonstrated significant within-subject differences in all three subjective outcomes of interest across the motion patterns. Post hoc pairwise comparison was conducted with the Holm-Bonferroni adjustment\cite{AGBANGBA2024e25131}. Among the four different behaviors of the robot tested in the \textit{Frontal Approach} scenario, the PLC behavior was evaluated by participants as significantly safer ($\bar{x}$=5.15, \textit{SD}=1.18) and smoother ($\bar{x}$=5.49, \textit{SD}=0.96) compared to all other tested behaviors. It was also evaluated as significantly more polite  ($\bar{x}$=5.43, \textit{SD}=0.98) compared to the slowing down and constant behaviors. The results appear to indicate that the motion pattern of PLC performed best according to the service objectives. Follow-up interviews indicated similar user perceptions, Participant 2 mentioned:
\say{When I could feel that the robot was recognizing me and taking avoidance action, I felt a sense of security, but when it didn't, I was worried about how it would move.}

%Also, Participant 6 added:%
%\say{I felt the politeness of not disturbing people when they came close to the wall as if to show from a distance that they were recognizing people.} %

The objective results, shown in \Cref{fig:eff_total score} were calculated based on the relative time it took to travel in each task compared to the user's control walking speed, denoted as efficiency. The PLC behavior did not result in higher efficiency of user movement based on this objective measure. However, some participants perceived that they could move the most efficiently with the PLC behavior.
Participant 33 mentioned:
\say{The robot moving away, means that I have more space for myself. Right? So, if it moves away, I can just keep going. So, this gives me a sense of efficiency. Maybe that's the reason why I don't like it stopping completely because it also kind of stops me.}
%Also, Participant 4 said:%
%\say{...because the robot's behaviors was the easiest to predict.}%

The results indicate that the answer to RQ1 is yes, the extended human detection capability by the PLC improves user impression towards the delivery robot compared to other behaviors. 

\subsection*{II. Blind Corner}
Referring again the subjective results, shown in \Cref{fig:result-individual}, unlike the \textit{Frontal Approach} scenario, a significant within-subject difference was found only for the safety category.

For the objective results, shown in \Cref{fig:eff_total score}, similar to \textit{Frontal Approach} no significant difference was found in terms of efficiency. The debriefing interviews conducted revealed that the perceived differences between the robot behaviors were very subtle, with many participants suggesting that they barely felt the difference in impression coming from the change in its motion.
Participant 4 mentioned:
\say{I did not feel there was a difference such as stopping or slowing down.
I only understood it when I saw the task diagram.
However, there were some differences, such as difficulty and ease of passing each other.}
Even for those who could feel the difference, their preferred behavior varied, thus the lack of significance in the intersection category.

The results indicate the answer to RQ2 is no, the social navigation system with PLC does not improve impressions when implemented in a residential setting with limited visibility.

\section{Discussions}
This paper contributes to the domain of socially compliant robot navigation systems in an indoor residential setup.
The presented Smart Logistics robot can react to people in an extended range and perform PLC behavior well in advance of the encounter.
This study was conducted to evaluate this PLC behavior with 42 participants across two scenarios: \textit{Frontal Approach} and \textit{Blind Corner} providing analysis from both quantitative and qualitative aspects.
Compared to typical short-range reactive social navigation approaches, the study revealed that participants preferred the PLC behavior significantly in accordance with Smart Logistics' service objectives in the \textit{Frontal Approach} scenario.
However, the analysis did not show any significant differences in the participants impression for the \textit{Blind Corner} scenario.
This appears to indicate that the navigation system with PLC does not necessarily improve user impression in parts of the hallway with limited visibility, where the advantages of long-range human sensing capability could not be realized.

As this study was conducted using autonomous robot motion, not all experimental values could be controlled precisely. Such as the exact location at which the robot was encountered in the \textit{Blind Corner} scenario, introducing limitations on the consistency of the results. This study also observed the influence of personal values toward non-humanoid robots and its effect on behavior preferences.
Some participants indicated that they preferred if the robot moved in smooth, continuous motions, not taking any preemptive action, and the participants themselves gave way to the robot.
Participant 7 mentioned:
\say{I kind of prefer no change, if we're just walking past and the robot is not doing anything special.}
Participant 16 added:
\say{When it turns, I don't want it to suddenly shift the left and turn, but rather to turn in a natural way.
I don't want it to suddenly come at me from the left after taking a wide turn to avoid humans.}
While others preferred clearly seeing signs that the robot recognized them by taking action, such as stopping, slowing down, or changing lanes, as Participant 40 mentioned:
\say{I think that the fact that the robot recognizes you and responds to you gives you a sense of security.}
From these responses, it was determined that both the smooth natural motion of the robot, and a sense of being detected/recognized by the robot were key requirements in feeling comfortable.

In the \textit{Frontal Approach} scenario, the PLC behavior could meet both requirements by taking action early enough to be predictable and clearly showing recognition by giving way with the lane-changing motion.
However, this was not the case in the \textit{Blind Corner} where no clear conclusion could be made about any of the behavior with regard to the requirements.

Referring to \Cref{fig:result-individual}, although the average scores for task B-$4'$ were lower than those for task B-4 in all categories (safety, smoothness, and politeness) for the \textit{Blind Corner} scenario, the lack of statistical significance ($p > 0.05$) indicates that this difference cannot be relied upon to draw definitive conclusions. A purely theoretical explanation may be that the PLC's consistent leftward lane changes, combined with the robot's tendency to cut corners during left turns, are associated with participants' feelings of discomfort. Further experiments would need to be done in order to verify this hypothesis.

The results show that it is challenging to provide an appropriate behavior in \textit{Blind Corners} solely relying on the motion of the robot.
Future studies are required to uncover if other approaches can be leveraged to improve the interactions, such as combining light and/or sound, which has been proven to impact perception on social behavior\cite{app14145983, MUTHUGALA2020103015}.
In addition, considering this delivery robot will be operating in residential buildings, future studies should address more diverse participants such as children, the elderly, and people with disabilities.

Future studies can also help to understand the generalizability of the presented approach and identify other domains or situations where robots might benefit from long-range social interaction.

Finally, our approach highlights the effectiveness of a mixed-method research approach (quantitative and qualitative).
The quantitative data helped clarify the validity of the PLC behavior statistically, and qualitative data identified the reasons why users scored behaviors more positively or negatively, ultimately helping to uncover more concrete solutions for further improvement.
Open-ended interviews provided valuable insights regarding user experience as per \citet{10.1145/1473018.1473055}.

\section*{Acknowledgments}

Thanks to the study participants and the Smart Logistics Robotics Development Team for having made this work possible. Also, we would like to thank May Woo for her advice on research design and analyses.

%% Use plainnat to work nicely with natbib.
\balance
\bibliographystyle{plainnat}
\bibliography{references}

@article{geldenbott2024legible,
  title   = {Legible and Proactive Robot Planning for Prosocial Human-Robot Interactions},
  author  = {Geldenbott, Jasper and Leung, Karen},
  journal = {arXiv preprint arXiv:2404.03734},
  year    = {2024}
}

@phdthesis{kirby2010social,
  title  = {Social robot navigation},
  author = {Kirby, Rachel},
  year   = {2010},
  month  = {June},
  school = {Carnegie Mellon University},
  type   = {PhD thesis},
  note   = {Available at \url{https://www.ri.cmu.edu/pub_files/2010/5/rk_thesis.pdf}}
}

@inproceedings{macenski2020marathon,
  title        = {The marathon 2: A navigation system},
  author       = {Macenski, Steve and Mart{\'\i}n, Francisco and White, Ruffin and Clavero, Jonatan Gin{\'e}s},
  booktitle    = {2020 IEEE/RSJ International Conference on Intelligent Robots and Systems (IROS)},
  pages        = {2718--2725},
  year         = {2020},
  organization = {IEEE}
}

@article{mavrogiannis2023core,
  title     = {Core challenges of social robot navigation: A survey},
  author    = {Mavrogiannis, Christoforos and Baldini, Francesca and Wang, Allan and Zhao, Dapeng and Trautman, Pete and Steinfeld, Aaron and Oh, Jean},
  journal   = {ACM Transactions on Human-Robot Interaction},
  volume    = {12},
  number    = {3},
  pages     = {1--39},
  year      = {2023},
  publisher = {ACM New York, NY}
}

@INPROCEEDINGS{5453230,
  author={Tsui, Katherine M. and Desai, Munjal and Yanco, Holly A.},
  booktitle={2010 5th ACM/IEEE International Conference on Human-Robot Interaction (HRI)}, 
  title={Considering the bystander's perspective for indirect human-robot interaction}, 
  year={2010},
  volume={},
  number={},
  pages={129-130},
  doi={10.1109/HRI.2010.5453230}}

@article{singh2023behavior,
  title={Behavior of Delivery Robot in Human-Robot Collaborative Spaces During Navigation.},
  author={Singh, Kiran Jot and Kapoor, Divneet Singh and Abouhawwash, Mohamed and Al-Amri, Jehad F and Mahajan, Shubham and Pandit, Amit Kant},
  journal={Intelligent Automation \& Soft Computing},
  volume={35},
  number={1},
  year={2023}
}

@misc{e-gov,
  url       = {https://laws.e-gov.go.jp/law/325CO0000000338/20230526_504CO0000000393#Mp-Ch_5-Se_2},
  author    = {Ministry of Land, Infrastructure, Transport and Tourism},
  journal   = {Building Standards Act of Japan},
  publisher = {Ministry of Internal Affairs and Communications},
  year      = {2023},
  note = {Accessed: 2024-09-13}
}

@Article{s20236822,
AUTHOR = {Guillén Ruiz, Silvia and Calderita, Luis V. and Hidalgo-Paniagua, Alejandro and Bandera Rubio, Juan P.},
TITLE = {Measuring Smoothness as a Factor for Efficient and Socially Accepted Robot Motion},
JOURNAL = {Sensors},
VOLUME = {20},
YEAR = {2020},
NUMBER = {23},
ARTICLE-NUMBER = {6822},
URL = {https://www.mdpi.com/1424-8220/20/23/6822},
PubMedID = {33260323},
ISSN = {1424-8220},
DOI = {10.3390/s20236822}
}

@inproceedings{lee2011effect,
  title={The effect of politeness strategy on human-robot collaborative interaction on malfunction of robot vacuum cleaner},
  author={Lee, Yeoreum and Bae, Jae-eul and Kwak, Sona S and Kim, Myung-Suk},
  booktitle={RSS workshop on HRI},
  year={2011}
}

@misc{ANOVA,
  url       = {https://www.datanovia.com/en/lessons/repeated-measures-anova-in-r/},
  author    = {Alboukadel Kassambara},
  publisher = {DATA NOVIA},
  year      = {2019},
  note = {Accessed: 2024-09-11}
}

@article{AGBANGBA2024e25131,
title = {On the use of post-hoc tests in environmental and biological sciences: A critical review},
journal = {Heliyon},
volume = {10},
number = {3},
pages = {e25131},
year = {2024},
issn = {2405-8440},
doi = {https://doi.org/10.1016/j.heliyon.2024.e25131},
url = {https://www.sciencedirect.com/science/article/pii/S2405844024011629},
author = {Codjo Emile Agbangba and Edmond Sacla Aide and Hermann Honfo and Romain Glèlè Kakai}
}

@misc{TWCweb,
  url       = {https://www.woven-city.global/},
  author    = {Toyota Woven City},
  journal   = {Toyota Woven City},
  publisher = {Woven by Toyota},
  year      = {2024}, 
  note = {Accessed: 2024-09-01}
}

@article{AKALIN2022102744,
title = {Do you feel safe with your robot? Factors influencing perceived safety in human-robot interaction based on subjective and objective measures},
journal = {International Journal of Human-Computer Studies},
volume = {158},
pages = {102744},
year = {2022},
issn = {1071-5819},
doi = {https://doi.org/10.1016/j.ijhcs.2021.102744},
url = {https://www.sciencedirect.com/science/article/pii/S1071581921001622},
author = {Neziha Akalin and Annica Kristoffersson and Amy Loutfi}
}

@InProceedings{xiao2022learning,
  title = 	 {Learning Model Predictive Controllers with Real-Time Attention for Real-World Navigation},
  author =       {Xiao, Xuesu and Zhang, Tingnan and Choromanski, Krzysztof Marcin and Lee, Tsang-Wei Edward and Francis, Anthony and Varley, Jake and Tu, Stephen and Singh, Sumeet and Xu, Peng and Xia, Fei and Persson, Sven Mikael and Kalashnikov, Dmitry and Takayama, Leila and Frostig, Roy and Tan, Jie and Parada, Carolina and Sindhwani, Vikas},
  booktitle = 	 {Proceedings of The 6th Conference on Robot Learning},
  pages = 	 {1708--1721},
  year = 	 {2023},
  editor = 	 {Liu, Karen and Kulic, Dana and Ichnowski, Jeff},
  volume = 	 {205},
  series = 	 {Proceedings of Machine Learning Research},
  month = 	 {14--18 Dec},
  publisher =    {PMLR},
  url = 	 {https://proceedings.mlr.press/v205/xiao23a.html}
}

@InProceedings{wang2022yolov7trainablebagoffreebiessets,
    author    = {Wang, Chien-Yao and Bochkovskiy, Alexey and Liao, Hong-Yuan Mark},
    title     = {YOLOv7: Trainable Bag-of-Freebies Sets New State-of-the-Art for Real-Time Object Detectors},
    booktitle = {Proceedings of the IEEE/CVF Conference on Computer Vision and Pattern Recognition (CVPR)},
    month     = {June},
    year      = {2023},
    pages     = {7464-7475}
}

@inproceedings{vaswani2023attentionneed,
 author = {Vaswani, Ashish and Shazeer, Noam and Parmar, Niki and Uszkoreit, Jakob and Jones, Llion and Gomez, Aidan N and Kaiser, \L ukasz and Polosukhin, Illia},
 booktitle = {Advances in Neural Information Processing Systems},
 editor = {I. Guyon and U. Von Luxburg and S. Bengio and H. Wallach and R. Fergus and S. Vishwanathan and R. Garnett},
 pages = {},
 publisher = {Curran Associates, Inc.},
 title = {Attention is All you Need},
 url = {https://proceedings.neurips.cc/paper_files/paper/2017/file/3f5ee243547dee91fbd053c1c4a845aa-Paper.pdf},
 volume = {30},
 year = {2017}
}

@inproceedings{gibert2013makes,
  title={What makes human so different? Analysis of human-humanoid robot interaction with a super Wizard of Oz platform},
  author={Gibert, Guillaume and Petit, Maxime and Lance, Florian and Pointeau, Gr{\'e}goire and Dominey, Peter Ford},
  booktitle={International Conference on Intelligent Robots and Systems},
  year={2013}
}

@INPROCEEDINGS{8172365,
  author={Law, Edith and Cai, Vicky and Liu, Qi Feng and Sasy, Sajin and Goh, Joslin and Blidaru, Alex and Kulić, Dana},
  booktitle={2017 26th IEEE International Symposium on Robot and Human Interactive Communication (RO-MAN)}, 
  title={A Wizard-of-Oz study of curiosity in human-robot interaction}, 
  year={2017},
  volume={},
  number={},
  pages={607-614},
  doi={10.1109/ROMAN.2017.8172365}
}

@INPROCEEDINGS{8172374,
  author={Tozadore, Daniel and Pinto, Adam and Romero, Roseli and Trovato, Gabriele},
  booktitle={2017 26th IEEE International Symposium on Robot and Human Interactive Communication (RO-MAN)}, 
  title={Wizard of Oz vs autonomous: Children's perception changes according to robot's operation condition}, 
  year={2017},
  volume={},
  number={},
  pages={664-669},
  doi={10.1109/ROMAN.2017.8172374}
}

@inproceedings{exploratory,
author = {Zamfirescu-Pereira, J.D. and Sirkin, David and Goedicke, David and LC, Ray and Friedman, Natalie and Mandel, Ilan and Martelaro, Nikolas and Ju, Wendy},
title = {Fake It to Make It: Exploratory Prototyping in HRI},
year = {2021},
isbn = {9781450382908},
publisher = {Association for Computing Machinery},
address = {New York, NY, USA},
url = {https://doi.org/10.1145/3434074.3446909},
doi = {10.1145/3434074.3446909},
booktitle = {Companion of the 2021 ACM/IEEE International Conference on Human-Robot Interaction},
pages = {19–28},
numpages = {10},
location = {Boulder, CO, USA},
series = {HRI '21 Companion}
}

@inproceedings{10.1145/2701973.2702060,
author = {Feng, Catherine and Azenkot, Shiri and Cakmak, Maya},
title = {Designing a Robot Guide for Blind People in Indoor Environments},
year = {2015},
isbn = {9781450333184},
publisher = {Association for Computing Machinery},
address = {New York, NY, USA},
url = {https://doi.org/10.1145/2701973.2702060},
doi = {10.1145/2701973.2702060},
booktitle = {Proceedings of the Tenth Annual ACM/IEEE International Conference on Human-Robot Interaction Extended Abstracts},
pages = {107–108},
numpages = {2},
location = {Portland, Oregon, USA},
series = {HRI'15 Extended Abstracts}
}

@inproceedings{10.1145/3461778.3462056,
author = {Kim, Keunwoo and Park, Minjung and Lim, Youn-kyung},
title = {Guiding Preferred Driving Style Using Voice in Autonomous Vehicles: An On-Road Wizard-of-Oz Study},
year = {2021},
isbn = {9781450384766},
publisher = {Association for Computing Machinery},
address = {New York, NY, USA},
url = {https://doi.org/10.1145/3461778.3462056},
doi = {10.1145/3461778.3462056},
booktitle = {Proceedings of the 2021 ACM Designing Interactive Systems Conference},
pages = {352–364},
numpages = {13},
location = {Virtual Event, USA},
series = {DIS '21}
}

@INPROCEEDINGS{8569486,
  author={Fuest, Tanja and Michalowski, Lars and Träris, Luca and Bellem, Hanna and Bengler, Klaus},
  booktitle={2018 21st International Conference on Intelligent Transportation Systems (ITSC)}, 
  title={Using the Driving Behavior of an Automated Vehicle to Communicate Intentions - A Wizard of Oz Study}, 
  year={2018},
  volume={},
  number={},
  pages={3596-3601},
  doi={10.1109/ITSC.2018.8569486}}

@conference{driving28548,
	author = {Brian Ka-Jun Mok and David Sirkin and Srinath Sibi and David Bryan Miller and Wendy Ju},
	title = {Understanding Driver-Automated Vehicle Interactions Through Wizard of Oz Design Improvisation},
	volume = {8},
	year = {2015},
	url = {https://pubs.lib.uiowa.edu/driving/article/id/28548/},
	issue = {2015},
	doi = {10.17077/drivingassessment.1598},
	month = {6},
	pages = {380-386},
	
	
	publisher={University of Iowa},
	journal = {Driving Assessment Conference}
}

@inproceedings{robotaxi_wooz,
author = {Meurer, Johanna and Pakusch, Christina and Stevens, Gunnar and Randall, Dave and Wulf, Volker},
title = {A Wizard of Oz Study on Passengers' Experiences of a Robo-Taxi Service in Real-Life Settings},
year = {2020},
isbn = {9781450369749},
publisher = {Association for Computing Machinery},
address = {New York, NY, USA},
url = {https://doi.org/10.1145/3357236.3395465},
doi = {10.1145/3357236.3395465},
booktitle = {Proceedings of the 2020 ACM Designing Interactive Systems Conference},
pages = {1365–1377},
numpages = {13},
location = {Eindhoven, Netherlands},
series = {DIS '20}
}

@article{7Likert,
  title={Relationships between ignoring instructions and response bias when completing questionnaires},
  author={Shinya Masuda and Takayuki Sakagami and Kazuyo Kitaoka and Megumi Sasaki},
  journal={The Japanese Journal of Psychology},
  volume={87},
  number={4},
  pages={354--363},
  year={2016},
  note={(in Japanese)}
}

@article{article_kumar,
author = {Shikhar Kumar and Eliran Itzhak and Yael Edan and Galit Nimrod and Vardit Sarne-Fleischmann and Noam Tractinsky},
year = {2022},
month = {08},
pages = {},
title = {Politeness in Human–Robot Interaction: A Multi-Experiment Study with Non-Humanoid Robots},
volume = {14},
journal = {International Journal of Social Robotics},
doi = {10.1007/s12369-022-00911-z}
}

@misc{NASA,
  url       = {https://spacecraft.ssl.umd.edu/design_lib/SSP50005rC.ISS_crew_integ.pdf},
  journal   = {International Space Station Flight
Crew Integration Standard
(NASA–STD–3000/T)},
  author = {National Aeronautics and Space Administration
Space Station Program Office
Johnson Space Center
Houston, Texas},
  year      = {1999},
  note = {Accessed: 2024-09-13}
}

@Article{app14145983,
AUTHOR = {Xu, Fan and Liu, Duanduan and Zhou, Chao and Hu, Jing},
TITLE = {Robotic Delivery Worker in the Dark: Assessment of Perceived Safety from Sidewalk Autonomous Delivery Robots’ Lighting Colors},
JOURNAL = {Applied Sciences},
VOLUME = {14},
YEAR = {2024},
NUMBER = {14},
ARTICLE-NUMBER = {5983},
URL = {https://www.mdpi.com/2076-3417/14/14/5983},
ISSN = {2076-3417},
DOI = {10.3390/app14145983}
}

@article{MUTHUGALA2020103015,
title = {Expressing attention requirement of a floor cleaning robot through interactive lights},
journal = {Automation in Construction},
volume = {110},
pages = {103015},
year = {2020},
issn = {0926-5805},
doi = {https://doi.org/10.1016/j.autcon.2019.103015},
url = {https://www.sciencedirect.com/science/article/pii/S0926580518313281},
author = {M. A. Viraj J. Muthugala and Ayyalusami Vengadesh and Xinke Wu and Mohan {Rajesh Elara} and Masami Iwase and Lingyun Sun and Jiang Hao}
}

@ARTICLE{DWA,
  author={Fox, D. and Burgard, W. and Thrun, S.},
  journal={IEEE Robotics \& Automation Magazine}, 
  title={The dynamic window approach to collision avoidance}, 
  year={1997},
  volume={4},
  number={1},
  pages={23-33},
  doi={10.1109/100.580977}}

@INPROCEEDINGS{MPPI_7487277,
  author={Williams, Grady and Drews, Paul and Goldfain, Brian and Rehg, James M. and Theodorou, Evangelos A.},
  booktitle={2016 IEEE International Conference on Robotics and Automation (ICRA)}, 
  title={Aggressive driving with model predictive path integral control}, 
  year={2016},
  volume={},
  number={},
  pages={1433-1440},
  doi={10.1109/ICRA.2016.7487277}}

@inproceedings{10.1145/1473018.1473055,
author = {Syrdal, Dag Sverre and Otero, Nuno and Dautenhahn, Kerstin},
title = {Video prototyping in human-robot interaction: results from a qualitative study},
year = {2008},
isbn = {9781605583990},
publisher = {Association for Computing Machinery},
address = {New York, NY, USA},
url = {https://doi.org/10.1145/1473018.1473055},
doi = {10.1145/1473018.1473055},
booktitle = {Proceedings of the 15th European Conference on Cognitive Ergonomics: The Ergonomics of Cool Interaction},
articleno = {29},
numpages = {8},
location = {Funchal, Portugal},
series = {ECCE '08}
}

@inproceedings{distance,
author = {Walters, Michael and Dautenhahn, Kerstin and Koay, Kheng and Kaouri, Christina and Boekhorst, Rene and Nehaniv, Chrystopher and Werry, Iain and Lee, D.},
year = {2005},
month = {02},
pages = {450 - 455},
title = {Close encounters: spatial distances between people and a robot of mechanistic appearance},
isbn = {0-7803-9320-1},
doi = {10.1109/ICHR.2005.1573608}
}

@InProceedings{10.1007/978-3-319-50115-4_69,
author="Dubois, Magda
and Claret, Josep-Arnau
and Basa{\~{n}}ez, Luis
and Venture, Gentiane",
editor="Kuli{\'{c}}, Dana
and Nakamura, Yoshihiko
and Khatib, Oussama
and Venture, Gentiane",
title="Influence of Emotional Motions in Human-Robot Interactions",
booktitle="2016 International Symposium on Experimental Robotics",
year="2017",
publisher="Springer International Publishing",
address="Cham",
pages="799--808"
}

@inproceedings{usenko2018double,
  title={The double sphere camera model},
  author={Usenko, Vladyslav and Demmel, Nikolaus and Cremers, Daniel},
  booktitle={2018 International Conference on 3D Vision (3DV)},
  pages={552--560},
  year={2018},
  organization={IEEE}
}

@inproceedings{kirillov2023segment,
  title={Segment anything},
  author={Kirillov, Alexander and Mintun, Eric and Ravi, Nikhila and Mao, Hanzi and Rolland, Chloe and Gustafson, Laura and Xiao, Tete and Whitehead, Spencer and Berg, Alexander C and Lo, Wan-Yen and others},
  booktitle={Proceedings of the IEEE/CVF International Conference on Computer Vision},
  pages={4015--4026},
  year={2023}
}

@misc{mmpose2020,
    title={OpenMMLab Pose Estimation Toolbox and Benchmark},
    author={MMPose Contributors},
    howpublished = {\url{https://github.com/open-mmlab/mmpose}},
    year={2020},
    note = {Accessed: 2024-09-15}
}

@article{LEICHTMANN2020101386,
title = {How much distance do humans keep toward robots? Literature review, meta-analysis, and theoretical considerations on personal space in human-robot interaction},
journal = {Journal of Environmental Psychology},
volume = {68},
pages = {101386},
year = {2020},
issn = {0272-4944},
doi = {https://doi.org/10.1016/j.jenvp.2019.101386},
url = {https://www.sciencedirect.com/science/article/pii/S0272494419303846},
author = {Benedikt Leichtmann and Verena Nitsch}
}

@INPROCEEDINGS{companion_5326271,
  author={Kirby, Rachel and Simmons, Reid and Forlizzi, Jodi},
  booktitle={RO-MAN 2009 - The 18th IEEE International Symposium on Robot and Human Interactive Communication}, 
  title={COMPANION: A Constraint-Optimizing Method for Person-Acceptable Navigation}, 
  year={2009},
  volume={},
  number={},
  pages={607-612},
  doi={10.1109/ROMAN.2009.5326271}}

@article{seitz2016cognitive,
  title={How cognitive heuristics can explain social interactions in spatial movement},
  author={Seitz, Michael J and Bode, Nikolai WF and K{\"o}ster, Gerta},
  journal={Journal of the Royal Society Interface},
  volume={13},
  number={121},
  pages={20160439},
  year={2016},
  publisher={The Royal Society}
}

@INPROCEEDINGS{physiological,
  author={Kivrak, Hasan and Uluer, Pinar and Kose, Hatice and Gumuslu, Elif and Erol Barkana, Duygun and Cakmak, Furkan and Yavuz, Sirma},
  booktitle={2020 29th IEEE International Conference on Robot and Human Interactive Communication (RO-MAN)}, 
  title={Physiological Data-Based Evaluation of a Social Robot Navigation System}, 
  year={2020},
  volume={},
  number={},
  pages={994-999},
  doi={10.1109/RO-MAN47096.2020.9223539}}

@inproceedings{shah2023vint,
  title     = {Vi{NT}: A Foundation Model for Visual Navigation},
  author    = {Dhruv Shah and Ajay Sridhar and Nitish Dashora and Kyle Stachowicz and Kevin Black and Noriaki Hirose and Sergey Levine},
  booktitle = {7th Annual Conference on Robot Learning},
  year      = {2023},
  url       = {https://arxiv.org/abs/2306.14846}
}

@article{kamezaki2020preliminary,
  title={A preliminary study of interactive navigation framework with situation-adaptive multimodal inducement: Pass-by scenario},
  author={Kamezaki, Mitsuhiro and Kobayashi, Ayano and Yokoyama, Yuta and Yanagawa, Hayato and Shrestha, Moondeep and Sugano, Shigeki},
  journal={International Journal of Social Robotics},
  volume={12},
  pages={567--588},
  year={2020},
  publisher={Springer}
}

@article{SFM,
  title={Social force model for pedestrian dynamics},
  author={Helbing, Dirk and Molnar, Peter},
  journal={Physical review E},
  volume={51},
  number={5},
  pages={4282},
  year={1995},
  publisher={APS}
}

@inproceedings{IRL_ziebart2009planning,
  title={Planning-based prediction for pedestrians},
  author={Ziebart, Brian D and Ratliff, Nathan and Gallagher, Garratt and Mertz, Christoph and Peterson, Kevin and Bagnell, J Andrew and Hebert, Martial and Dey, Anind K and Srinivasa, Siddhartha},
  booktitle={2009 IEEE/RSJ International Conference on Intelligent Robots and Systems},
  pages={3931--3936},
  year={2009},
  organization={IEEE}
}

@inproceedings{IRL_henry2010learning,
  title={Learning to navigate through crowded environments},
  author={Henry, Peter and Vollmer, Christian and Ferris, Brian and Fox, Dieter},
  booktitle={2010 IEEE international conference on robotics and automation},
  pages={981--986},
  year={2010},
  organization={IEEE}
}

@article{IRL_kretzschmar2016socially,
  title={Socially compliant mobile robot navigation via inverse reinforcement learning},
  author={Kretzschmar, Henrik and Spies, Markus and Sprunk, Christoph and Burgard, Wolfram},
  journal={The International Journal of Robotics Research},
  volume={35},
  number={11},
  pages={1289--1307},
  year={2016},
  publisher={SAGE Publications Sage UK: London, England}
}

@inproceedings{ORCA_van2011reciprocal,
  title={Reciprocal n-body collision avoidance},
  author={Van Den Berg, Jur and Guy, Stephen J and Lin, Ming and Manocha, Dinesh},
  booktitle={Robotics Research: The 14th International Symposium ISRR},
  pages={3--19},
  year={2011},
  organization={Springer}
}

@article{Francis2023PrinciplesAG,
  title={Principles and Guidelines for Evaluating Social Robot Navigation Algorithms},
  author={Anthony Francis and Claudia P{\'e}rez-D'Arpino and Chengshu Li and Fei Xia and Alexandre Alahi and Rachid Alami and Aniket Bera and Abhijat Biswas and Joydeep Biswas and Rohan Chandra and Hao-Tien Lewis Chiang and Michael Everett and Sehoon Ha and Justin W. Hart and Jonathan P. How and Haresh Karnan and Tsang-Wei Edward Lee and Luis J. Manso and Reuth Mirksy and Soeren Pirk and Phani Teja Singamaneni and Peter Stone and Ada V Taylor and Pete Trautman and Nathan Tsoi and Marynel V{\'a}zquez and Xuesu Xiao and Peng Xu and Naoki Yokoyama and Alexander Toshev and Roberto Martin-Martin},
  journal={ArXiv},
  year={2023},
  volume={abs/2306.16740},
  url={https://api.semanticscholar.org/CorpusID:259287246}
}

@article{article_Likert,
author = {Joshi, Ankur and Kale, Saket and Chandel, Satish and Pal, Dinesh},
year = {2015},
month = {01},
pages = {396-403},
title = {Likert Scale: Explored and Explained},
volume = {7},
journal = {British Journal of Applied Science \& Technology},
doi = {10.9734/BJAST/2015/14975}
}

@inproceedings{edirisinghe2024field,
  title={Field Trial of an Autonomous Shopworker Robot that Aims to Provide Friendly Encouragement and Exert Social Pressure},
  author={Edirisinghe, Sachi and Satake, Satoru and Brscic, Drazen and Liu, Yuyi and Kanda, Takayuki},
  booktitle={Proceedings of the 2024 ACM/IEEE International Conference on Human-Robot Interaction},
  pages={194--202},
  year={2024}
}

@article{hellou2022technical,
  title={Technical methods for social robots in museum settings: An overview of the literature},
  author={Hellou, Mehdi and Lim, JongYoon and Gasteiger, Norina and Jang, Minsu and Ahn, Ho Seok},
  journal={International Journal of Social Robotics},
  volume={14},
  number={8},
  pages={1767--1786},
  year={2022},
  publisher={Springer}
}

@INPROCEEDINGS{TEB,
  author={Roesmann, Christoph and Feiten, Wendelin and Woesch, Thomas and Hoffmann, Frank and Bertram, Torsten},
  booktitle={ROBOTIK 2012; 7th German Conference on Robotics}, 
  title={Trajectory modification considering dynamic constraints of autonomous robots}, 
  year={2012},
  volume={},
  number={},
  pages={1-6},
  doi={}}

@inproceedings{person_detector,
author = {Jia, Dan and Steinweg, Mats and Hermans, Alexander and Leibe, Bastian},
title = {Self-Supervised Person Detection in 2D Range Data using a Calibrated Camera},
year = {2021},
publisher = {IEEE Press},
url = {https://doi.org/10.1109/ICRA48506.2021.9561699},
doi = {10.1109/ICRA48506.2021.9561699},
booktitle = {2021 IEEE International Conference on Robotics and Automation (ICRA)},
pages = {13301–13307},
numpages = {7},
location = {Xi'an, China}
}

@article{hasan_lidar-based_2022,
	title = {{LiDAR}-based detection, tracking, and property estimation: A contemporary review},
	volume = {506},
	issn = {09252312},
	url = {https://linkinghub.elsevier.com/retrieve/pii/S0925231222009365},
	doi = {10.1016/j.neucom.2022.07.087},
	shorttitle = {{LiDAR}-based detection, tracking, and property estimation},
	pages = {393--405},
	journaltitle = {Neurocomputing},
	shortjournal = {Neurocomputing},
	author = {Hasan, Mahmudul and Hanawa, Junichi and Goto, Riku and Suzuki, Ryota and Fukuda, Hisato and Kuno, Yoshinori and Kobayashi, Yoshinori},
	urldate = {2024-11-11},
	date = {2022-09},
	langid = {english},
}

\end{document}